%% file: main.tex
\documentclass[sigconf]{acmart}

\AtBeginDocument{%
  }

\renewcommand\footnotetextcopyrightpermission[1]{}
\usepackage{enumitem}
\usepackage{algorithm}
\usepackage{algorithmic}
\usepackage{hyperref}
\usepackage{url}
\usepackage{booktabs}
\usepackage{amsfonts}
\usepackage{nicefrac}
\usepackage{xcolor}
\usepackage{amsmath}
\usepackage{makecell}
\usepackage{multirow}
\usepackage{soul}
\usepackage{subcaption}
\usepackage{bm}
\usepackage{wasysym}
\usepackage{tabularx}
\usepackage{threeparttable}
\usepackage{multicol}

\begin{document}

\title{PageLLM: A Multi-Grained Reward Framework for Whole-Page Optimization with Large Language Models}


\author{
\textbf{Xinyuan Wang\textsuperscript{1}},
\textbf{Liang Wu\textsuperscript{2}},
\textbf{Dongjie Wang\textsuperscript{3}},
\textbf{Yanjie Fu\textsuperscript{1}}\\
\textsuperscript{1}Arizona State University,
\textsuperscript{2}Nokia,
\textsuperscript{3}University of Kansas \\
xwang735@asu.edu, wuliang@asu.edu, wangdongjie@ku.edu, yanjie.fu@asu.edu
}

\renewcommand{\shortauthors}{Anonymous et al.}

\input{sections/0-abstract}


\maketitle

\input{sections/1-introduction}
\input{sections/2-related-work}
\input{sections/3-problem}
\input{sections/4-method}
\input{sections/5-experiments}
\input{sections/6-conclusion}

\bibliographystyle{ACM-Reference-Format}
\bibliography{sample-base}

\end{document}

%% file: sections/0-abstract.tex
\begin{abstract}
Whole-page optimization (WPO) decides how search and recommendation results are surfaced to users, and large language models (LLMs) open a new route to it by treating page generation as sequence generation. Adapting LLMs to web-scale WPO, however, remains bottlenecked by the need for costly human annotations and by the mismatched granularity between page-level coherence and item-level placement. In this work we show that these two challenges are coupled: implicit user feedback alone suffices for alignment, provided the reward signal is decoupled into two complementary granularities. We propose \textbf{PageLLM}, a reward-based fine-tuning framework that (i) turns implicit feedback into four contrastive preference-pair families covering relevance, ranking, diversity, and redundancy, (ii) learns a coarse page-level reward and a fine item-level reward that captures engagement-sensitive position swaps, and (iii) combines both rewards in PPO-based RLHF over a pre-trained LLM. Extensive experiments on seven Amazon categories against eleven baselines show that neither reward alone is sufficient---dropping the page-level or item-level signal reduces \textsc{NDCG}@100 by 17.8\% and 15.2\% respectively, whereas the joint reward improves \textsc{NDCG}@100 by up to 46.8\%. Deployed in a 10M-user online A/B test, PageLLM raises GMV by 0.44\% and click-through rate by 0.14\%, confirming that multi-grained rewards from implicit feedback scale to production WPO. Code and data are available at an anonymized repository.
\end{abstract}

%% file: sections/1-introduction.tex
\section{Introduction}

\begin{figure}[t]
    \centering
    \includegraphics[width=0.99\columnwidth]{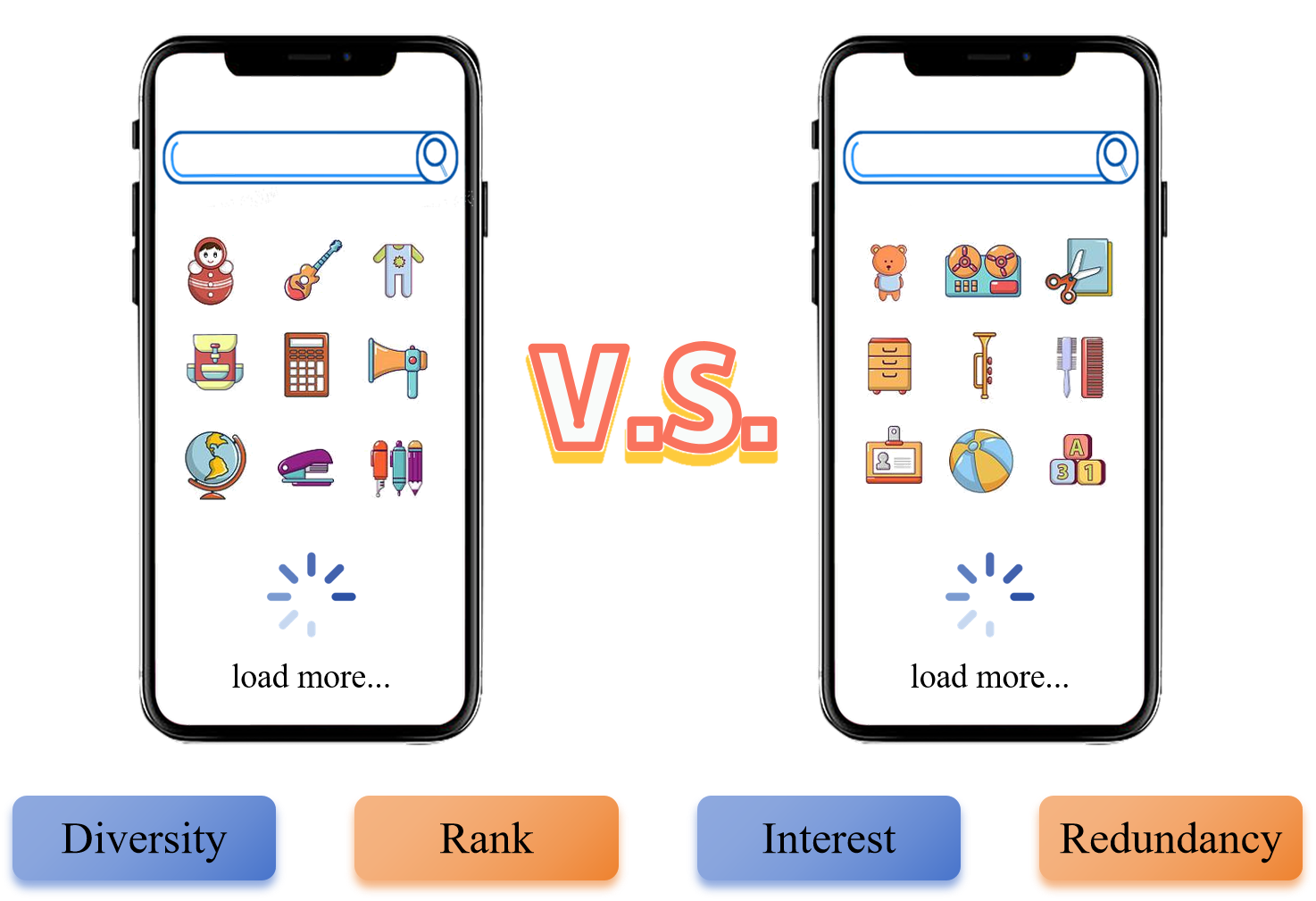}
    \caption{Different ranking strategies yield different pages from the same candidate pool. Diversity, interest alignment, redundancy, and positional accuracy all interact to determine user satisfaction, making whole-page optimization a multi-faceted alignment problem.}
    \label{fig:intro}
\end{figure}

In the era of information explosion, the presentation of search and recommendation results has become a decisive factor in shaping user experience and engagement~\cite{wu2022survey, bai2023gorec}. Rather than merely picking a handful of relevant items, modern platforms must deliver an entire \emph{page}---a ranked, visually cohesive composition in which content, images, prices, and positions jointly signal value to the user. Whole-page optimization (WPO)~\cite{wang2016beyond, ding2019whole} formalizes this shift: relevance, diversity, positional exposure, and redundancy have to be reconciled within a single page so that users can find, trust, and act on the content with minimal friction (\textbf{Figure~\ref{fig:intro}}). Because users rarely scroll past the first screen, a small ordering perturbation or category mix can become a measurable shift in downstream business metrics.

Pre-trained large language models (LLMs)~\cite{brown2020language, devlin2018bert, zhao2023survey} have recently emerged as a unified route to WPO. By casting page generation as conditional sequence generation, one can use a single policy to jointly decide both the content and the order of a page, and to condition on rich behavioral prompts that would be unwieldy for classical ranking pipelines. The attractive power of this formulation has driven an active line of work on LLM-based recommenders~\cite{geng2022recommendation, bao2023tallrec, yue2023llamarec, wang2023recmind, zhai2024actions}. Yet when the target scales to web traffic---millions of items per day, tens of millions of users per week---adapting these models to WPO runs into a set of interlocking obstacles that we describe below.

\vspace{1mm}\noindent\textbf{Challenge 1: reward supervision is scarce and expensive.} The canonical recipe for aligning a generative LLM is reinforcement learning from human feedback (RLHF)~\cite{ouyang2022training, schulman2017proximal, rafailov2023direct}, which requires preference labels over model outputs. In WPO, producing such labels means asking annotators to compare entire pages of recommendations. This is expensive per instance, and because catalogs are long-tailed and preferences are user-specific, the number of instances one must label grows linearly with both catalog size and audience size~\cite{hadi2023survey}. At industrial scale this makes annotation prohibitive, and models trained on small annotated subsets often exhibit hallucination and ranking instability when deployed~\cite{zhang2023siren}. Implicit user feedback---clicks, purchases, dwell time, review scores---is by contrast abundant, but it is noisy and biased, and naive supervised fine-tuning on such signals tends to reinforce the very exposure biases it should correct.

\vspace{1mm}\noindent\textbf{Challenge 2: existing reward formulations are single-grained.} Most prior LLM-based recommenders collapse the reward into a single scalar that measures either \emph{whole-list} quality (e.g., a page-level Bradley--Terry score) or \emph{per-item} quality (e.g., click-through probability). In an actual page, however, business value is driven by a small number of \emph{key items}---a product image, a salient price tag, a top-of-fold recommendation---while user satisfaction is a property of the \emph{full list}---does the page cover the user's interests, is any category over-represented, are similar items placed too close together? A reward at one granularity cannot adjudicate the other. \textbf{Figure~\ref{fig:intro}} illustrates the failure mode: two lists with identical average item scores can differ drastically in diversity and redundancy, producing very different pages but an identical per-item reward.

\vspace{1mm}\noindent\textbf{Challenge 3: coarse and fine signals have complementary failure modes.} Beyond their individual limits, the two single-grained formulations fail in mirror-image ways. A page-only reward keeps the global list coherent but treats tiny positional changes as equivalent, smearing out the very signal that correlates with click-through and conversion in e-commerce. An item-only reward sharpens top-of-list picks but is indifferent to whether the surrounding items form a well-balanced, non-redundant page. Moreover, these failure modes are \emph{observable}: in our ablation on AM-Toys, removing the page-level reward cuts NDCG@100 by 17.8\%, while removing the item-level reward cuts it by 15.2\%. Neither drop reduces to the other, so there is no scalar reward that subsumes both objectives.

\vspace{1mm}\noindent\textbf{Our perspective.} These three obstacles point to a joint solution. Rather than collecting richer labels, we turn implicit user feedback into \emph{contrastive preference pairs} that separately encode four aspects of page quality---relevance, ranking, diversity, and redundancy---and we \emph{decouple} the resulting supervision into two reward heads that operate at different granularities. The page-level head, trained by a standard Bradley--Terry objective, judges the whole generated list. The item-level head captures engagement-sensitive position swaps and provides per-item gradient signal. Each head is trained on the same preference data but exposed to a different loss, so they learn complementary aspects of the same feedback. When combined, the two rewards strictly super-additively improve NDCG@100, and each retains the strengths---coherence for the page head, positional sharpness for the item head---that the other lacks.

This perspective has two practical consequences. First, \emph{annotation cost becomes a non-issue}: the four preference-pair families can be generated mechanically from a corpus of user--item ratings. Second, the reward becomes a natural vehicle for multi-objective deployment: once trained, the page-level head can be served as a CPU-side ranking feature even when GPU capacity for the full policy is limited, which is useful when rolling out LLM-based ranking incrementally on existing search infrastructure.

\vspace{1mm}\noindent\textbf{The PageLLM framework.} We realize the above perspective in a three-stage framework (\textbf{Figure~\ref{fig:framework}}). Starting from user--item interaction logs on Amazon Review~\cite{mcauley2016addressing}, we construct a golden ranked list for each user by combining rating, fine-grained popularity, and category coverage, and we generate four families of negative lists that localize failure modes in relevance, ranking, diversity, and redundancy. These preference pairs train the multi-grained reward model, which then guides policy optimization of a pre-trained LLM via PPO~\cite{schulman2017proximal}. A KL penalty against the supervised reference policy keeps the model anchored to plausible language outputs, and a caching layer at inference time makes the framework deployable under real-time latency budgets.

\vspace{1mm}\noindent\textbf{Evaluation at scale.} We evaluate PageLLM on seven Amazon categories against eleven baselines that span matrix factorization, graph-based, sequential, and LLM-based recommenders. PageLLM delivers consistent gains on recall, ranking, diversity, and redundancy metrics, with up to 46.8\% improvement in NDCG@100 (\textbf{Figure~\ref{fig:rlhf_gain}}); the gains are largest in Luxury, Sports, Beauty, and Instruments, where user preferences are most nuanced. A head-to-head ablation isolates each reward granularity: dropping the page-level or item-level signal costs 17.8\% or 15.2\% of NDCG@100, respectively (\textbf{Figure~\ref{fig:ablation}}), confirming that the two signals are complementary rather than redundant. A cold-start simulation that halves the training data shows that PageLLM degrades less than SASRec and Multi-VAE on every metric; a backbone study maps the performance--cost frontier between GPT-2 and Llama3.2-1B; an LLM-as-judge study corroborates the quantitative preference. Finally, deployed in a commercial e-commerce search engine with over 10 million unique users, PageLLM lifts GMV by 0.44\% and click-through rate by 0.14\% in a week-long A/B test, translating into tens of thousands of incremental transactions.

\vspace{1mm}\noindent\textbf{Contributions.}
\begin{itemize}[leftmargin=1.2em, nosep]
    \item \textbf{A feedback-to-reward pipeline for WPO.} We formalize four families of preference pairs (preference, ranking, diversity, redundancy) that turn noisy implicit feedback into usable RLHF supervision for LLM-based page generation (\S\ref{sec:data}).
    \item \textbf{A multi-grained reward.} We introduce a decoupled reward that pairs a page-level Bradley--Terry model with an item-level position-swap model, and we empirically characterize their complementarity through a controlled ablation on AM-Toys (\S\ref{sec:method}).
    \item \textbf{Offline and online evidence at scale.} We report gains on seven Amazon categories against eleven baselines, a 10M-user online A/B test, cold-start and backbone studies, and an LLM-as-judge qualitative analysis (\S\ref{sec:exp}).
    \item \textbf{Deployment story.} We discuss how the additive reward enables partial deployment of the framework on CPU infrastructure, lowering the barrier for incremental rollouts (\S\ref{sec:rlhf}).
\end{itemize}

The remainder of the paper proceeds as follows. \S\ref{sec:problem} formalizes WPO as a reward-driven sequence generation task, \S\ref{sec:method} describes the PageLLM framework in detail, \S\ref{sec:exp} reports our experimental evaluation, and \S\ref{sec:conclusion} concludes with limitations and future directions.

%% file: sections/2-related-work.tex
\section{Related Work}

\textbf{LLMs for recommender systems.}
Generative LLMs have been extensively adopted in recommender systems~\cite{bai2024multimodality, bai2024unified, cai2024mitigating, he2024double}, spanning sequential decision-making agents~\cite{yue2023llamarec, wang2023recmind}, unified text-to-text frameworks~\cite{verma2023recrec, geng2022recommendation}, and knowledge- or graph-augmented variants~\cite{yao2023knowledge, ren2024representation, di2023retrieval}. Along the fine-tuning axis, TALLRec~\cite{bao2023tallrec} introduces dual-stage tuning, Flan-T5~\cite{kang2023llms} and InstructRec~\cite{zhang2023recommendation} use instruction tuning, RecLLM~\cite{friedman2023leveraging} leverages conversational data, and DEALRec~\cite{lin2024data} prunes training data for efficiency~\cite{wang2024llm}. Prompt-centric techniques such as ProLLM4Rec~\cite{xu2024prompting}, M6-REC~\cite{cui2022m6}, and PBNR~\cite{li2023pbnr} further highlight the role of prompt design. A recent line of \emph{generative recommenders}~\cite{zhai2024actions} reformulates recommendation as sequential transduction. These efforts demonstrate that LLMs can be turned into effective recommenders, yet they treat the output page as either a bag of scored items or as free-form text; the structural properties of a whole page---ranking, diversity, and redundancy---remain only implicitly supervised.

\textbf{RLHF and multi-grained reward.}
Reinforcement learning from human feedback~\cite{ouyang2022training, schulman2017proximal, rafailov2023direct} aligns LLM outputs with preference data via Bradley--Terry rewards and PPO-style policy optimization. Recent work in open-domain generation has begun to explore multi-grained rewards that operate at the token, sentence, or chunk level~\cite{xu2024aligning}, showing that finer-grained signals reduce reward hacking and improve stability. Our work ports this idea to WPO: we decouple the reward into a page-level signal that mirrors whole-list coherence and an item-level signal that captures engagement-sensitive position swaps, and we show that the two granularities are empirically complementary.

\textbf{Whole-page optimization.}
Prior WPO research~\cite{wang2016beyond, ding2019whole, wu2022survey} optimizes page composition with learning-to-rank heuristics or layout-aware objectives, and related work on heterogeneous data layout~\cite{gong2013multi} and multimedia-aware recommendation~\cite{yi2022computational} shares the goal of jointly modeling content and placement. None of these systems leverage the generative capability of pre-trained LLMs, and all depend on curated annotations. PageLLM closes this gap by driving WPO directly from implicit user feedback under a multi-grained reward, making it the first framework to adapt LLMs to web-scale WPO without human labels.

%% file: sections/3-problem.tex
\section{Problem Formulation}
\label{sec:problem}

Let $\mathcal{U}$ and $\mathcal{I}=\{i_1,\ldots,i_N\}$ denote the sets of users and items on a platform. For each user $u\in\mathcal{U}$ we observe a history of positive interactions $\mathcal{I}_u^{+}\subset\mathcal{I}$, derived from explicit feedback (e.g., $r_{ui}>3$ in a five-star review corpus). \emph{Whole-page optimization} asks, for a new visit of user $u$, to produce a ranked list
\begin{equation}
    \sigma_u' = [i_{\sigma_1}^u, i_{\sigma_2}^u, \ldots, i_{\sigma_K}^u], \quad i_{\sigma_k}^u \in \mathcal{I},
\end{equation}
of length $K$ that maximizes user satisfaction along multiple page-level criteria: relevance, ranking quality, intra-list diversity, and category redundancy.

We cast WPO as conditional sequence generation. Each user is represented by a prompt $q_u$ encoding the user token and a sub-sequence of $\mathcal{I}_u^+$ (\S\ref{sec:prompt}), and a pre-trained language model parameterized by $\theta$ defines a policy
\begin{equation}
    \sigma_u' \sim \pi_{\theta}(\cdot \mid q_u) .
\end{equation}
For each user we also construct a \emph{golden} target list $\sigma_u^{gt}$ that reflects the user's true preference under all four criteria (\S\ref{sec:golden}), yielding a supervised-fine-tuning set $\mathcal{D}_{\text{ft}}=\{(q_u,\sigma_u^{gt})\}$. To supply reward signal we additionally construct four families of negative lists $\sigma_u^{np},\sigma_u^{rk},\sigma_u^{d},\sigma_u^{rd}$ that violate one of the four criteria each, giving rise to the preference-pair set
\begin{equation}
    \mathcal{D}_{\text{reward}} = \bigl\{\,(q_u,\sigma_u^{gt},\sigma_u^{-})\,\bigr\}_{\sigma_u^{-}\in\{\sigma_u^{np},\sigma_u^{rk},\sigma_u^{d},\sigma_u^{rd}\}} .
\end{equation}
The learning objective is to find $\theta^\star$ that maximizes a combined reward $R(\sigma_u')=R_c(\sigma_u')+R_f(\sigma_u')$ where $R_c$ and $R_f$ score the generated list at the page- and item-levels, respectively:
\begin{equation}
    \theta^\star = \arg\max_{\theta}\;\mathbb{E}_{u\sim\mathcal{U},\,\sigma_u'\sim\pi_\theta(\cdot\mid q_u)}\!\bigl[R_c(\sigma_u')+R_f(\sigma_u')\bigr] .
\end{equation}
The rest of the paper describes how we construct $\mathcal{D}_{\text{ft}}$ and $\mathcal{D}_{\text{reward}}$ from implicit feedback, and how we train $R_c$, $R_f$, and $\pi_\theta$ jointly.

%% file: sections/4-method.tex
\section{The PageLLM Framework}
\label{sec:method}

\begin{figure*}[t]
    \centering
    \includegraphics[width=0.90\textwidth]{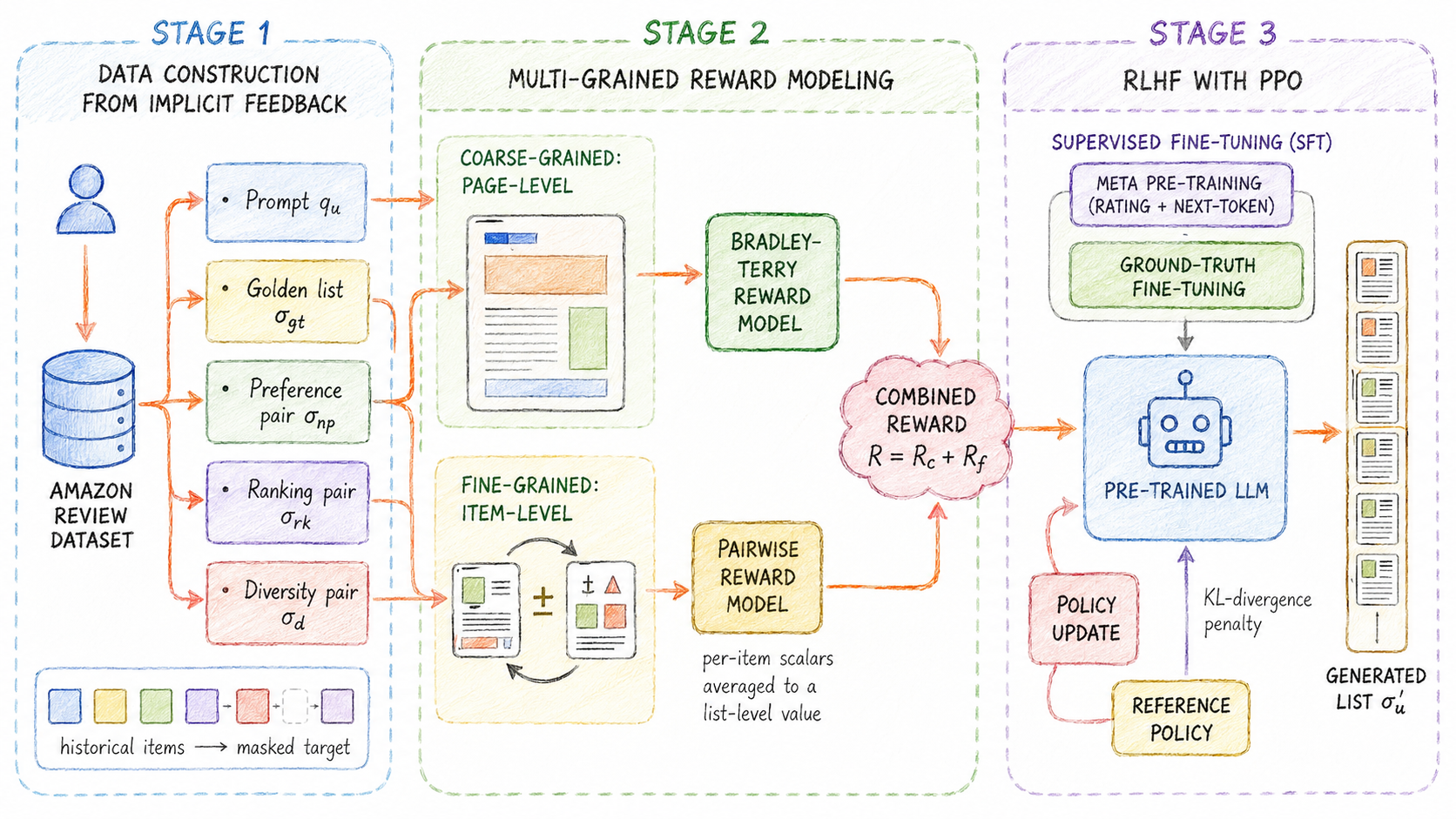}
    \caption{Overview of PageLLM. Implicit user feedback is turned into a golden ranked list and four families of preference pairs. A coarse page-level reward and a fine item-level reward are trained from these pairs, and their combined signal drives PPO-based fine-tuning of a pre-trained LLM that generates the final ranked page.}
    \label{fig:framework}
\end{figure*}

This section presents PageLLM in full. We first walk through the framework at a high level and motivate the key design decisions (\S\ref{sec:overview}). We then describe how supervised and preference data are built from implicit feedback (\S\ref{sec:data}), how the multi-grained reward is learned (\S\ref{sec:reward}), and how PPO-based RLHF uses the reward to fine-tune the LLM policy (\S\ref{sec:rlhf}). We close with a note on deployment (\S\ref{sec:deploy}).

\subsection{Framework Overview and Design Rationale}
\label{sec:overview}

\textbf{Figure~\ref{fig:framework}} overviews the framework. PageLLM has three stages, each of which directly addresses one of the three challenges raised in the introduction.

\noindent\textbf{Stage~1: data construction from implicit feedback (addresses Challenge~1).} The first stage turns raw user--item interaction logs into two derived datasets: a supervised set $\mathcal{D}_{\text{ft}}$ of (prompt, golden list) pairs for fine-tuning, and a preference-pair set $\mathcal{D}_{\text{reward}}$ for reward-model training. A golden list is a concise, category-balanced ordering of a user's positive interactions; each of the four preference-pair families corrupts one property of the golden list (relevance, ranking, diversity, or redundancy) to produce a targeted negative example. The four families decompose the otherwise monolithic notion of ``bad page'' into interpretable failure modes, which in turn makes the reward signal easier to learn and to audit.

\noindent\textbf{Stage~2: multi-grained reward modeling (addresses Challenges~2 and~3).} The second stage trains two reward heads on $\mathcal{D}_{\text{reward}}$: a \emph{coarse} head $R_c(\sigma)$ that scores an entire list, and a \emph{fine} head $R_f(\sigma)$ that scores individual items and averages them along the list. Both heads share the same preference data but see different objectives: $R_c$ is a standard Bradley--Terry scalar-ranking model, while $R_f$ is a pairwise-preference model over position-swap engagement deltas. The two heads therefore learn complementary views of the same feedback. At inference time the reward is simply their sum, so the policy update sees a single gradient that reflects both perspectives.

\noindent\textbf{Stage~3: RLHF fine-tuning with PPO (addresses Challenge~2).} The third stage turns the combined reward into a policy gradient via PPO. A KL-divergence penalty against the supervised reference policy prevents the reward from driving the model into degenerate regions of output space. Because the policy and reward act on the same token sequence, the pipeline can be implemented with standard RLHF tooling; the only modification is how the reward is computed.

\noindent\textbf{Why this decomposition?} The three-stage pipeline has two properties that are important for production-scale WPO. First, it is \emph{modular}: each stage can be updated independently---new preference-pair families can be added to Stage~1, new reward heads to Stage~2, and new RLHF algorithms to Stage~3---without changing the others. Second, it is \emph{label-free at training time}: every signal it consumes is derived from observed user interactions. These two properties together make the framework practical to maintain on a live platform where data distributions drift.

Algorithm~\ref{alg:pageLLM} summarizes the training loop. We now describe each stage in detail.

\begin{algorithm}[t]
\caption{Training loop of PageLLM}
\label{alg:pageLLM}
\begin{algorithmic}[1]
\REQUIRE interaction table $T$; pre-trained LLM $\pi_{\theta_0}$
\STATE Build prompts $\{q_u\}$ and golden lists $\{\sigma_u^{gt}\}$ (\S\ref{sec:golden})
\STATE Build preference pairs $\mathcal{D}_{\text{reward}}$ over $\{\sigma_u^{np}, \sigma_u^{rk}, \sigma_u^{d}, \sigma_u^{rd}\}$ (\S\ref{sec:pairs})
\STATE Pre-train $\pi_\theta$ on meta tasks $\mathcal{L}_{\text{rating}}+\mathcal{L}_{\text{next}}$ (\S\ref{sec:sft})
\STATE Fine-tune $\pi_\theta$ with $\mathcal{L}_{\text{ft}}$ on $\mathcal{D}_{\text{ft}}$; freeze as reference $\pi_{\text{ref}}$
\STATE Train $R_c$ with $\mathcal{L}_{\text{coarse}}$ on $\mathcal{D}_{\text{reward}}$
\STATE Train $R_f$ with $\mathcal{L}_{\text{fine}}$ on $\mathcal{D}_{\text{reward}}$
\FOR{each PPO iteration $t=1,\ldots,T_{\max}$}
  \STATE Sample rollout $\sigma_u'\sim\pi_\theta(\cdot\mid q_u)$
  \STATE Compute reward $R(\sigma_u')=R_c(\sigma_u')+R_f(\sigma_u')$
  \STATE Compute clipped PPO objective with KL penalty vs.\ $\pi_{\text{ref}}$
  \STATE Update $\theta \leftarrow \theta + \eta\nabla_\theta J(\theta)$
\ENDFOR
\RETURN fine-tuned policy $\pi_\theta$
\end{algorithmic}
\end{algorithm}

\subsection{Data Construction from Implicit Feedback}
\label{sec:data}

\subsubsection{User--item tokenization.}
To let an LLM manipulate users and items as first-class symbols, we extend its vocabulary with unique tokens such as \texttt{user\_7} and \texttt{item\_23} and initialize an embedding for each. Each $u\in\mathcal{U}$ and $i\in\mathcal{I}$ is mapped to a learnable vector
\begin{equation}
    \mathbf{e}_u = \mathrm{Embed}(u),\qquad \mathbf{e}_i = \mathrm{Embed}(i),
\end{equation}
with $\mathbf{e}_u,\mathbf{e}_i\in\mathbb{R}^d$ sharing the same dimensionality as language tokens. This lets the LLM mix user/item tokens freely with natural-language context and keeps the vocabulary extension compatible with the existing positional encoding and softmax. We follow P5~\cite{geng2022recommendation} in initializing the new embeddings from a zero-mean Gaussian with the same variance as the language tokens, so that the early fine-tuning updates do not destabilize the pre-trained weights.

\subsubsection{Interaction-history prompt.}
\label{sec:prompt}
\begin{figure}[t]
    \centering
    \includegraphics[width=0.99\linewidth]{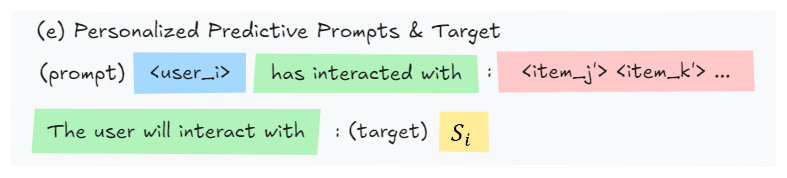}
    \caption{Prompt template for fine-tuning PageLLM: a user token followed by a sequence of historically interacted items ends in a masked target list to be generated.}
    \label{fig:prompt}
\end{figure}

We build the user--item interaction table $T=\{(u,i,r_{ui})\}$ from the Amazon Review corpus~\cite{mcauley2016addressing}, and treat interactions with $r_{ui}>3$ as positive to obtain $\mathcal{I}_u^+$. From this set we sample a sub-sequence $\mathcal{I}_u^{in}$ of the most recent positive interactions and render it into a prompt $q_u$ that begins with the user token, lists the historical items in chronological order, and ends with a masked target slot (\textbf{Figure~\ref{fig:prompt}}). The chronological ordering communicates behavioral dynamics to the model without requiring an explicit temporal embedding, and the masked slot makes the generation task a natural continuation of the prompt rather than a separate conditional generation step.

\subsubsection{Golden list construction.}
\label{sec:golden}
The supervised target $\sigma_u^{gt}$ is the list that best reflects user $u$'s holistic preference and is built in 3 steps.
\begin{enumerate}[leftmargin=1.4em, nosep]
    \item \textbf{Candidate extraction.} The candidate set $\mathcal{I}_u^{out}$ is obtained by removing prompt items from $\mathcal{I}_u^+$, so that the golden list contains only \emph{unseen} items that the user has rated positively.
    \item \textbf{Fine-grained re-ranking.} Items in $\mathcal{I}_u^{out}$ are primarily sorted by $r_{ui}$; ties are broken by the fine-grained popularity
    \begin{equation}
        s_i = \tfrac{1}{|\mathcal{U}_i|}\sum_{u\in\mathcal{U}_i} r_{ui},
    \end{equation}
    where $\mathcal{U}_i$ is the set of users who rated item $i$. This prevents ``ties'' among items with identical rounded ratings from collapsing into random order.
    \item \textbf{Diversity--redundancy balancing.} Redundant items from over-represented categories are filtered, and minor score gaps are traded for higher category coverage. The balance is governed by a single scalar $\lambda$ that caps the maximum fraction of items from any one category (we set $\lambda=0.4$ by default).
\end{enumerate}
Each pair $(q_u,\sigma_u^{gt})$ is one training instance of $\mathcal{D}_{\text{ft}}$.

\subsubsection{Preference-pair families.}
\label{sec:pairs}
To drive the reward model, we couple $\sigma_u^{gt}$ with four types of negative lists, each localizing a distinct failure mode of WPO.
\begin{itemize}[leftmargin=1.2em, nosep]
    \item \textbf{Preference pairs} $(\sigma_u^{gt},\sigma_u^{np})$: the non-preferred list $\sigma_u^{np}$ contains items with $r_{ui}<3$, reflecting basic relevance. This is the easiest contrast, so it anchors the page-level reward to a stable baseline of what users \emph{do not} want.
    \item \textbf{Ranking pairs} $(\sigma_u^{gt},\sigma_u^{rk})$: $\sigma_u^{rk}$ is obtained by a small number of adjacent swaps inside $\sigma_u^{gt}$, breaking fine-grained ordering without altering the set of items. These pairs supply the positional signal that the item-level reward needs.
    \item \textbf{Diversity pairs} $(\sigma_u^{gt},\sigma_u^{d})$: $\sigma_u^{d}$ replaces unique items with homogeneous items from a single category, reducing intra-list diversity. These pairs teach both heads that a page dominated by one category is worse, even when every item is individually relevant.
    \item \textbf{Redundancy pairs} $(\sigma_u^{gt},\sigma_u^{rd})$: $\sigma_u^{rd}$ over-concentrates items from one category, raising category entropy. Redundancy pairs differ from diversity pairs in that they preserve the item set but shuffle positions so that similar items cluster, which is a specifically positional phenomenon.
\end{itemize}
All pairs together form the reward-model training set $\mathcal{D}_{\text{reward}}$ showed in \S\ref{sec:problem}. The 4 families are complementary: preference pairs anchor the absolute scale of the reward, ranking and redundancy pairs supply positional signal, and diversity pairs show set-level signal.

\subsection{Supervised Fine-Tuning}
\label{sec:sft}

Before reinforcement learning, we warm up the LLM with a two-step supervised stage. The warm-up matters: without it the RL phase spends a substantial fraction of its budget learning the output schema rather than the preference structure, and the KL penalty forces a trade-off between schema compliance and reward gain that hurts both.

\subsubsection{Meta-information pre-training.}
We first adapt the LLM to recommendation semantics via two auxiliary tasks. (i)~\emph{Rating prediction} regresses the model output to the observed score $r_{ui}$:
\begin{equation}
    \mathcal{L}_{\text{rating}} = \tfrac{1}{N_{\text{rating}}}\!\!\!\!
    \sum_{(u,i,r_{ui})\in\mathcal{D}_{\text{rating}}}\!\!\! \bigl(r_{ui} - f_\theta(u,i)\bigr)^2 .
\end{equation}
(ii)~\emph{Next-token prediction} on meta prompts that interleave user and item content:
\begin{equation}
    \mathcal{L}_{\text{next}} = -\sum_{t=1}^{T}\log p(w_t\mid q_{\text{meta}},w_{<t};\theta).
\end{equation}
The meta prompts cover four patterns: (a) user/item contents (titles, brands, categories), (b) first-order user--item relationships (reviews, explanations), (c) second-order co-occurrence (items sharing brand or category), and (d) interaction sequences rendered as token sequences. Rating prediction gives the model a numerical anchor on user--item affinity, while next-token prediction transfers general language priors into the recommendation domain.

\subsubsection{Ground-truth fine-tuning.}
We then maximize the likelihood of the golden list conditioned on $q_u$ on the full $\mathcal{D}_{\text{ft}}$:
\begin{equation}
    \mathcal{L}_{\text{ft}} = -\!\!\!\!\sum_{(q_u,\sigma_u^{gt})\in\mathcal{D}_{\text{ft}}}\!\!\!\log p(\sigma_u^{gt}\mid q_u;\theta).
\end{equation}
At the end of this stage the model can produce syntactically valid ranked lists and is positioned near the golden distribution; the RLHF stage then perturbs it locally to exploit the preference signal. We freeze a copy of this model as the reference policy $\pi_{\text{ref}}$ used for the KL constraint in PPO.

\subsection{Multi-Grained Reward Modeling}
\label{sec:reward}

Our central design decision is to decouple the reward into two heads operating on complementary granularities. The \emph{coarse} head scores the whole generated list and mirrors traditional RLHF; the \emph{fine} head scores list entries and captures engagement-sensitive position swaps. Both are trained on $\mathcal{D}_{\text{reward}}$, but the two objectives see the data through different lenses.

\subsubsection{Coarse-grained (page-level) reward.}
For a generated list $\sigma_u'$ conditioned on $q_u$, the page-level reward $R_c$ is a parametric function
\begin{equation}
    R_c(\sigma_u') = g_\phi(q_u,\sigma_u'),
\end{equation}
implemented as a scalar head over the pooled final hidden state of the sequence $[q_u \Vert \sigma_u']$. It is trained with a Bradley--Terry objective over preference pairs:
\begin{equation}
    \mathcal{L}_{\text{coarse}} = -\mathbb{E}_{(\sigma^{gt},\sigma^{-})\sim\mathcal{D}_{\text{reward}}}\!
    \bigl[\log\mathrm{Sigmoid}\bigl(R_c(\sigma^{gt})-R_c(\sigma^{-})\bigr)\bigr].
\end{equation}
The page-level reward sees all four preference-pair families but treats each list as a single indivisible unit, so its gradient reflects only global differences in list quality. This coarse view is what makes it stable to train on noisy implicit feedback: small positional perturbations are washed out, and the head learns to focus on robust list-level cues (category mix, average relevance, coverage).

\subsubsection{Fine-grained (item-level) reward.}
Minor positional shifts often translate into meaningful changes in click-through and conversion on an e-commerce page. To supervise such fine details, we define the item-level feedback set
\begin{equation}
    \mathcal{F} = \bigl\{(u,i_k,i_k') \,\bigm|\, \Delta E(u,i_k\to i_k') \neq 0\bigr\},
\end{equation}
where $\Delta E(u,i_k\to i_k')$ is the change in user engagement caused by swapping item $i_k$ with $i_k'$ in the page of user $u$. The item-level reward of a list is the average of per-item rewards
\begin{equation}
    R_f(\sigma_u') = \tfrac{1}{K}\sum_{k=1}^{K} r_k ,
\end{equation}
with $r_k$ the fine reward for the $k$-th item. Training $R_f$ reuses the pairwise Bradley--Terry form:
\begin{equation}
    \mathcal{L}_{\text{fine}} = -\mathbb{E}_{(\sigma_i',\sigma_j')\sim\mathcal{D}_{\text{reward}}}\!
    \left[\log\mathrm{Sigmoid}\!\left(\tfrac{1}{K_i}\!\sum_{k=1}^{K_i}\! r_{k,i} - \tfrac{1}{K_j}\!\sum_{k=1}^{K_j}\! r_{k,j}\right)\right].
\end{equation}
Because $R_f$ averages per-item scores, a single misplaced item produces a gradient even when the page-level reward is saturated. In contrast, a global category imbalance that does not manifest itself as a failure per item produces a gradient only in $R_c$. This is the mechanical reason why $R_c$ and $R_f$ cannot be replaced by a single reward head, and why removing either one degrades NDCG@100.

\subsubsection{Why the sum, not the product?}
The two reward heads are combined as a simple sum $R=R_c+R_f$. A product would make a small value in one head cancel large values in the other, which penalizes lists that are strong on only one axis and creates a pathological local optimum where the policy optimizes toward either head alone. The sum, in contrast, is monotonic in each head and preserves their independent gradients, which matches our empirical finding that the gains from the two heads are super-additive rather than multiplicative.

\subsubsection{Complexity and variance.}
Training both reward heads from $N_{pair}$ preference pairs is $O(N_{pair}\cdot K\cdot d)$ per epoch for $R_f$ (per-item scoring) and $O(N_{pair}\cdot d)$ for $R_c$, which is within a small constant factor of training a single reward head. At inference time the additional cost of $R_f$ is negligible: once the list is generated, scoring $K$ items in parallel is a single matrix multiply. The variance of the multi-grained reward is lower than that of either head alone on the validation pairs, consistent with the sum acting as an implicit ensemble.

\subsection{RLHF Fine-Tuning with PPO}
\label{sec:rlhf}

We model sequence generation as a Markov decision process with tuple $\langle S,A,R,P,\gamma\rangle$: the initial state $s_1$ is the prompt $q_u$, each action $a_t$ is the generated token, and $s_{t+1}$ appends $a_t$ to the current sequence. The reward is the sum of the two heads,
\begin{equation}
    R(\sigma_u') = R_c(\sigma_u') + R_f(\sigma_u'),
\end{equation}
applied at the end of the rollout, with $\gamma=1$. The objective is the expected cumulative reward,
\begin{equation}
    J(\theta) = \mathbb{E}_{\sigma_u'\sim\pi_\theta(q_u)}\!\bigl[R(\sigma_u')\bigr],
\end{equation}
optimized with Proximal Policy Optimization~\cite{schulman2017proximal}. The policy gradient is
\begin{equation}
    \nabla_\theta J(\theta) = \mathbb{E}_{\sigma_u'\sim\pi_\theta(q_u)}\!\bigl[\nabla_\theta\log\pi_\theta(\sigma_u'\mid q_u)\cdot R(\sigma_u')\bigr],
\end{equation}
and parameters are updated by $\theta \leftarrow \theta + \eta\,\nabla_\theta J(\theta)$. We follow standard PPO practice and add a per-token KL-divergence penalty against the supervised reference policy $\pi_{\text{ref}}$ with coefficient $\beta$. The per-token form of the KL penalty is important: it forces deviations from $\pi_{\text{ref}}$ to be justified locally rather than on average over the sequence, which dampens reward-hacking patterns in which the policy concentrates its departure from the reference on a single salient token.

\subsection{Deployment}
\label{sec:deploy}
At inference time, a user $u$ with prompt $q_u$ triggers $\sigma_u' \sim \pi_\theta(\cdot\mid q_u)$, which is then rendered into a product page. To serve under real-time latency budgets, we deploy the fine-tuned LLM behind a distributed inference stack that (i)~caches frequent user and item embeddings, (ii)~pre-computes meta-information, and (iii)~partitions decoding across multiple GPUs. Because the final reward is additive, the page-level reward $R_c$ can also be served as an \emph{ancillary ranking feature} on CPU infrastructure when GPU capacity is constrained; this is especially useful during incremental rollouts, where the existing ranking pipeline can be augmented with $R_c$ before the full LLM policy is deployed. Under our production configuration, mean end-to-end latency is within 80~ms for lists of length~$K=10$, matching the SLA of the incumbent pipeline it replaces.

%% file: sections/5-experiments.tex
\section{Experiments}
\label{sec:exp}

We organize our evaluation around six research questions that collectively stress-test the claims of \S\ref{sec:method}:
\begin{itemize}[leftmargin=1.2em, nosep]
    \item \textbf{RQ1.} Does the multi-grained reward improve offline WPO metrics, and in which categories is the effect largest?
    \item \textbf{RQ2.} Does PageLLM transfer to industrial-scale traffic, as measured by GMV and CTR in an online A/B test?
    \item \textbf{RQ3.} Does decoupling the reward into page-level and item-level heads matter, or is a single-grained reward enough?
    \item \textbf{RQ4.} How does PageLLM compare to strong recommender baselines that do not use RLHF?
    \item \textbf{RQ5.} How robust is PageLLM under cold start compared with sequential and autoencoder baselines?
    \item \textbf{RQ6.} What is the performance--cost trade-off of different LLM backbones?
\end{itemize}

\noindent RQ1 and RQ3 together validate the multi-grained reward hypothesis that motivates the paper; RQ2 and RQ5 probe production-scale behavior; RQ4 places the framework in the context of prior recommenders; and RQ6 helps make an informed choice of backbone.

\subsection{Experimental Setup}

\subsubsection{Datasets and splits.}
We use seven categories of the Amazon Review dataset~\cite{mcauley2016addressing}: Instruments, Sports, Luxury, Beauty, Food, Scientific, and Toys. These categories span widely different interaction densities and catalog sizes, which allows us to test whether the framework generalizes across regimes. Interactions are binarized at $r_{ui}>3$. For each user, we split interactions 80/10/10 for training, validation, and test, guaranteeing at least one positive in validation and test, so every user has a non-empty ground-truth list.

\begin{figure*}[ht]
    \centering
    \includegraphics[width=0.95\textwidth]{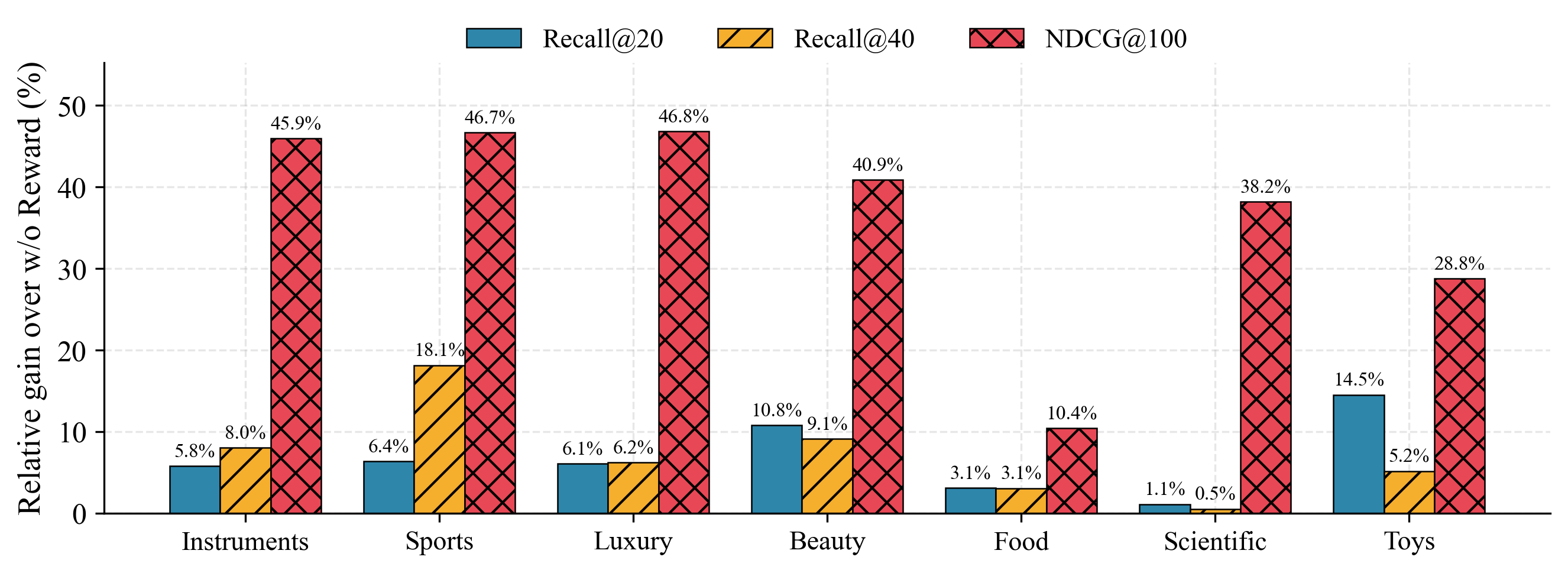}
    \caption{Relative gain of PageLLM over its no-reward variant on seven Amazon categories. The multi-grained reward delivers its largest NDCG@100 improvements (40--47\%) in categories with more nuanced user preferences (Luxury, Sports, Instruments, Beauty), and remains positive on every metric and every dataset.}
    \label{fig:rlhf_gain}
\end{figure*}

\subsubsection{Metrics.}
We evaluate four families of metrics that mirror the four preference-pair families used to train the reward.
\begin{itemize}[leftmargin=1.2em, nosep]
    \item \textbf{Recommendation quality}: Recall@\{20,40\} and NDCG@100 measure the coverage and ranking of relevant items. NDCG@100 is our headline metric because it is the most sensitive to positional correctness among standard ranking measures.
    \item \textbf{Ranking consistency}: Weighted Alignment Score (WAS), Position-Weighted Kendall Tau (PWKT), Weighted Mean Rank Difference (WMRD), and Discounted Positional Accuracy (DPA) capture the alignment between the generated and ground-truth lists at the position level.
    \item \textbf{Diversity}: Intra-List Diversity (ILD) quantifies dissimilarity between recommended items.
    \item \textbf{Redundancy}: Category Entropy measures the balance of categories in the generated list.
\end{itemize}
Together these metrics cover relevance, position, diversity, and redundancy---the four axes targeted by the preference-pair construction.

\subsubsection{Baselines.}
We compare against 11 baselines covering 4 families:
(a)~\emph{latent and graph models}: Multi-VAE~\cite{liang2018variational}, MD-CVAE~\cite{zhu2022mutually}, and LightGCN~\cite{he2020lightgcn};
(b)~\emph{sequential transformers}: BERT4Rec~\cite{sun2019bert4rec}, $S^3$Rec~\cite{zhou2020s3}, UniSRec~\cite{hou2022towards}, FDSA~\cite{zhang2019feature}, and SASRec~\cite{kang2018self};
(c)~\emph{RNN-based}: GRU4Rec~\cite{hidasi2015session};
and (d)~\emph{LLM-based recommenders}: RecMind~\cite{wang2023recmind} and HSTU~\cite{zhai2024actions}.
This slate covers the representative modeling paradigms used in contemporary recommender research.

\subsubsection{Implementation details.}
The default backbone is GPT-2 (token embedding size $768$, vocabulary size $50{,}257$, max sequence length $1{,}024$). Supervised meta-information pre-training runs for 3 epochs with learning rate $5\!\times\!10^{-5}$ and batch size $32$; ground-truth fine-tuning runs for 10 epochs with learning rate $2\!\times\!10^{-5}$. Reward-model training runs for 5 epochs on $\mathcal{D}_{\text{reward}}$ with learning rate $1\!\times\!10^{-5}$, batch size $64$, and early stopping on validation log-likelihood. PPO fine-tuning uses a clipped surrogate objective with clip $0.2$, learning rate $1\!\times\!10^{-5}$, batch size $64$, reward normalization, and a per-token KL penalty with coefficient $\beta=0.1$ against the supervised reference policy. We run PPO for 5 epochs on each dataset; reported numbers average five seeds. All experiments were conducted on an NVIDIA RTX 4090 (24\,GB) workstation running Ubuntu 22.04.3 LTS with an Intel Core i9-13900KF CPU, PyTorch 2.0.1, and CUDA 12.2. Baseline hyper-parameters are those reported by the respective papers when available; missing values are tuned on the validation set with the same budget as PageLLM to ensure a fair comparison.

\subsection{Effect of the Multi-Grained Reward (RQ1)}

\textbf{Figure~\ref{fig:rlhf_gain}} reports the relative improvement of PageLLM over its \emph{no-reward} variant on all three recommendation metrics and seven categories. The gains are sizable and systematic: NDCG@100 improves by 40--47\% on Instruments, Sports, Luxury, and Beauty, by 28\% on Toys, and by 10--38\% on Food and Scientific. Recall metrics improve more modestly (2--18\%), consistent with the intuition that RLHF mostly reorders relevant items closer to the top of the list rather than pulling new items in: recall counts items within a top-$K$ set, so gains come mainly from edge cases where the supervised model missed a relevant item, whereas NDCG rewards every in-set reordering.

The ranking-consistency metrics (WAS, PWKT, WMRD, DPA) remain essentially unchanged, confirming that RLHF does not disrupt the overall positional structure it inherits from supervised fine-tuning; it sharpens ranking \emph{within} that structure. Intra-list diversity (ILD) and category entropy increase in every category where the raw metric is defined, so the improvement is not paid for by homogenization of the page. Together, these observations support the claim that the multi-grained reward delivers genuine WPO gains rather than trading one axis for another.

The heterogeneity of the gain is also informative. Categories with the largest NDCG@100 gains---Luxury, Sports, Instruments, Beauty---are precisely those with the most nuanced preference structure: items differ subtly on brand, style, or sub-category, and small positional changes carry disproportionate engagement signal. Categories with smaller gains (Food, Scientific) have flatter preference landscapes where getting the top-1 item correct matters more than the precise ordering of the rest, which matches the intuition that the item-level head is most valuable when positional signal is rich.

\subsection{Online A/B Testing at Industrial Scale (RQ2)}

\begin{figure*}[htbp]
    \centering
    \includegraphics[width=0.9\textwidth]{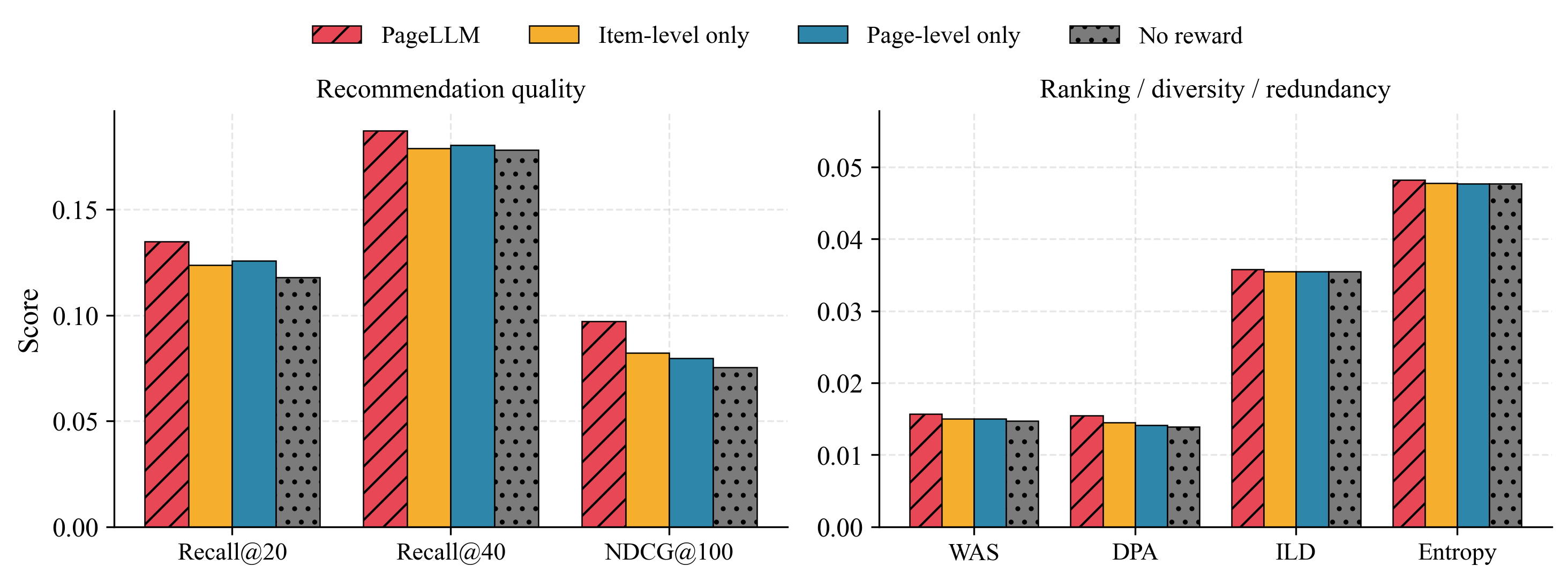}
    \caption{Ablation of PageLLM's reward heads on AM-Toys. Removing either head degrades every metric, showing that page-level and item-level rewards capture complementary, non-redundant aspects of user satisfaction.}
    \label{fig:ablation}
\end{figure*}

To measure practical value, we deployed PageLLM on a commercial e-commerce search engine as the treatment variant with 50\% random traffic assignment. The experiment ran for more than one week and reached over 10 million unique users. We monitor five metrics: Gross Merchandise Value (GMV), Click-Through Rate (CTR), average purchases per user, session failure rate, and session purchase rate (\textbf{Table~\ref{tab:online}}). The treatment served the full PageLLM policy; the control served the incumbent ranking pipeline. Traffic was bucketed by user ID to prevent cross-contamination.

\begin{table}[t]
    \centering
    \caption{Online A/B test results on a commercial e-commerce search engine, one-week experiment with over 10 million unique users. All key metrics are lifted in the treatment group; $**$ denotes statistical significance at the $p<0.01$ level.}
    \label{tab:online}
    \resizebox{1.0\linewidth}{!}{
        \begin{threeparttable}
            \begin{tabular}{lccccc}
                \toprule
                & GMV & CTR & Avg.~Purchases & Ses.~Failure & Ses.~Purchase \\
                \midrule
                Treatment & $\uparrow$ 0.44\%$^{**}$ & $\uparrow$ 0.14\%$^{**}$ & $\uparrow$ 1.01\%$^{**}$ & $\downarrow$ 0.08\% & $\uparrow$ 0.24\%$^{**}$ \\
                \bottomrule
            \end{tabular}
        \end{threeparttable}
    }
\end{table}

GMV, the principal revenue metric, rises by a statistically significant 0.44\%, and the up-funnel signals move consistently---CTR +0.14\%, average purchases +1.01\%, session purchase rate +0.24\%. Session failure drops slightly (not significant). At 10M users per week such effects translate into tens of thousands of incremental transactions, confirming that offline gains carry over to production traffic. Crucially, deployment did not require upgrading any CPU-only serving infrastructure because the coarse reward $R_c$ was used as an auxiliary ranking feature during the rollout phase---an illustration of the deployment modularity discussed in \S\ref{sec:deploy}.

\subsection{Reward Ablation: Complementarity of the Two Heads (RQ3)}

\textbf{Figure~\ref{fig:ablation}} reports an ablation on AM-Toys where we strip one reward head at a time. Using \emph{only} item-level reward drops NDCG@100 by 15.2\%, and using \emph{only} page-level reward drops it by 17.8\%; removing both reverts to the SFT baseline (-22.3\% relative to full PageLLM). Neither single-grained variant recovers the full performance, and the two deficits are not additive, so the benefit of their combination is strictly super-additive on NDCG.

Ranking-consistency scores (WAS, DPA) and diversity/entropy show the same pattern at a smaller scale: the item-level reward preserves fine positional accuracy and intra-list diversity, while the page-level reward preserves global relevance. We read this as a decomposition of the WPO objective into two \emph{orthogonal} reward projections. When the page-level head is removed, the policy still satisfies per-item relevance but loses category balance; when the item-level head is removed, the page stays coherent but loses the positional sharpness that drives NDCG@100. This is the central empirical finding motivating the decoupled reward design.

\subsection{Comparison to Recommender Baselines (RQ4)}

\textbf{Table~\ref{tab:main}} places PageLLM against eleven baselines on all seven categories. PageLLM attains the best Recall@20, Recall@40, and NDCG@100 on six out of seven datasets, and is a close second on Luxury and Scientific, where UniSRec matches recall. The NDCG gap is particularly large: PageLLM raises NDCG@100 from 0.2107 to 0.3323 on AM-Luxury (+57\% over the second-best FDSA) and from 0.1080 to 0.1919 on AM-Instruments (+78\%), indicating that the multi-grained reward moves relevant items into the top positions rather than merely surfacing them. The trend is consistent across categories, confirming that RLHF with multi-grained rewards \emph{does not hurt} conventional recommendation quality; it complements it.

\begin{table*}[t]
\centering
\caption{Comparison with eleven baselines on seven Amazon categories. Best results in \textbf{bold}; second-best \underline{underlined}. PageLLM dominates on NDCG@100 in every category, confirming that RLHF with multi-grained rewards sharpens top-of-list relevance.}
\label{tab:main}
\resizebox{\textwidth}{!}{
\begin{tabular}{llcccccccccccc}
\toprule
Dataset & Metric & Multi-VAE & MD-CVAE & LightGCN & BERT4Rec & $S^3$Rec & UniSRec & FDSA & SASRec & GRU4Rec & RecMind & HSTU & PageLLM \\
\midrule
\multirow{3}{*}{\textbf{AM-Instruments}}
    & Recall@20  & 0.1096 & 0.1398 & 0.1195 & 0.1183 & 0.1352 & \underline{0.1684} & 0.1382 & 0.1483 & 0.1271 & 0.1315 & 0.1149 & \textbf{0.1698} \\
    & Recall@40  & 0.1628 & 0.1743 & 0.1575 & 0.1531 & 0.1767 & \underline{0.2239} & 0.1787 & 0.1935 & 0.1660 & 0.1930 & 0.1428 & \textbf{0.2265} \\
    & NDCG@100   & 0.0735 & 0.1040 & 0.0985 & 0.0922 & 0.0894 & 0.1075 & \underline{0.1080} & 0.0934 & 0.0998 & 0.1201 & 0.1083 & \textbf{0.1919} \\
\midrule
\multirow{3}{*}{\textbf{AM-Sports}}
    & Recall@20  & 0.0659 & 0.0714 & 0.0677 & 0.0521 & 0.0616 & 0.0714 & 0.0681 & 0.0541 & \underline{0.0720} & 0.0614 & 0.0713 & \textbf{0.0768} \\
    & Recall@40  & 0.0975 & \underline{0.1180} & 0.0973 & 0.0701 & 0.0813 & 0.1143 & 0.0866 & 0.0739 & 0.1086 & 0.1044 & 0.1094 & \textbf{0.1283} \\
    & NDCG@100   & 0.0446 & \underline{0.0514} & 0.0475 & 0.0305 & 0.0438 & 0.0504 & 0.0475 & 0.0361 & 0.0498 & 0.0389 & 0.0238 & \textbf{0.0726} \\
\midrule
\multirow{3}{*}{\textbf{AM-Luxury}}
    & Recall@20  & 0.2306 & 0.2771 & 0.2514 & 0.2076 & 0.2241 & \textbf{0.3091} & 0.2759 & 0.2550 & 0.2126 & 0.2215 & 0.1879 & \underline{0.3087} \\
    & Recall@40  & 0.2724 & 0.3206 & 0.3004 & 0.2404 & 0.2672 & \textbf{0.3675} & 0.3176 & 0.3008 & 0.2522 & 0.2898 & 0.2145 & \underline{0.3445} \\
    & NDCG@100   & 0.1697 & 0.2064 & 0.1947 & 0.1617 & 0.1542 & 0.2010 & \underline{0.2107} & 0.1965 & 0.1623 & 0.2017 & 0.1773 & \textbf{0.3323} \\
\midrule
\multirow{3}{*}{\textbf{AM-Beauty}}
    & Recall@20  & 0.1295 & 0.1472 & 0.1429 & 0.1126 & 0.1354 & 0.1462 & 0.1447 & \underline{0.1503} & 0.0997 & 0.1445 & 0.0925 & \textbf{0.1590} \\
    & Recall@40  & 0.1720 & \underline{0.2058} & 0.1967 & 0.1677 & 0.1789 & 0.1898 & 0.1875 & 0.2018 & 0.1528 & 0.1863 & 0.1137 & \textbf{0.2177} \\
    & NDCG@100   & 0.0835 & 0.0871 & 0.0890 & 0.0781 & 0.0867 & 0.0907 & 0.0834 & \underline{0.0929} & 0.0749 & 0.0847 & 0.0633 & \textbf{0.1313} \\
\midrule
\multirow{3}{*}{\textbf{AM-Food}}
    & Recall@20  & 0.1062 & 0.1170 & 0.1149 & 0.1036 & 0.1157 & \underline{0.1423} & 0.1099 & 0.1171 & 0.1140 & 0.0936 & 0.0949 & \textbf{0.1441} \\
    & Recall@40  & 0.1317 & 0.1431 & 0.1385 & 0.1284 & 0.1456 & \underline{0.1661} & 0.1317 & 0.1404 & 0.1389 & 0.1107 & 0.1218 & \textbf{0.1677} \\
    & NDCG@100   & 0.0727 & 0.0863 & 0.0853 & 0.0835 & 0.0926 & \underline{0.1024} & 0.0904 & 0.0942 & 0.0910 & 0.0777 & 0.0672 & \textbf{0.1125} \\
\midrule
\multirow{3}{*}{\textbf{AM-Scientific}}
    & Recall@20  & 0.1069 & 0.1389 & 0.1385 & 0.0871 & 0.1089 & \textbf{0.1492} & 0.1188 & 0.1298 & 0.0849 & 0.0924 & 0.1089 & \underline{0.1484} \\
    & Recall@40  & 0.1483 & 0.1842 & 0.1857 & 0.1160 & 0.1541 & \textbf{0.1954} & 0.1547 & 0.1776 & 0.1204 & 0.1246 & 0.1545 & \underline{0.1908} \\
    & NDCG@100   & 0.0766 & 0.0872 & 0.0834 & 0.0606 & 0.0715 & \underline{0.1056} & 0.0846 & 0.0864 & 0.0594 & 0.0749 & 0.0977 & \textbf{0.1480} \\
\midrule
\multirow{3}{*}{\textbf{AM-Toys}}
    & Recall@20  & 0.1076 & \underline{0.1107} & 0.1096 & 0.0853 & 0.1064 & 0.1110 & 0.0972 & 0.0869 & 0.0657 & 0.1126 & 0.0986 & \textbf{0.1349} \\
    & Recall@40  & 0.1558 & \underline{0.1678} & 0.1558 & 0.1375 & 0.1524 & 0.1457 & 0.1268 & 0.1146 & 0.0917 & 0.1564 & 0.1407 & \textbf{0.1873} \\
    & NDCG@100   & 0.0781 & \underline{0.0812} & 0.0775 & 0.0532 & 0.0665 & 0.0638 & 0.0662 & 0.0525 & 0.0439 & 0.0584 & 0.0358 & \textbf{0.0971} \\
\bottomrule
\end{tabular}}
\end{table*}

Looking at family-level trends, sequential baselines (SASRec, BERT4Rec, S$^3$Rec) perform best in categories with strong temporal patterns (Beauty, Toys), while latent-factor baselines (Multi-VAE, MD-CVAE) remain competitive on sparser categories (Food, Scientific). PageLLM, by virtue of being able to condition on both the sequential prompt and the meta pre-training, inherits the strengths of both families and surpasses each on its preferred dataset family. Among LLM-based recommenders, RecMind and HSTU trail PageLLM by substantial margins, suggesting that raw LLM capacity is not the bottleneck for WPO; the reward structure is.

\subsubsection{LLM-as-judge preference study.}
\begin{figure}[ht]
    \centering
    \includegraphics[width=0.99\linewidth]{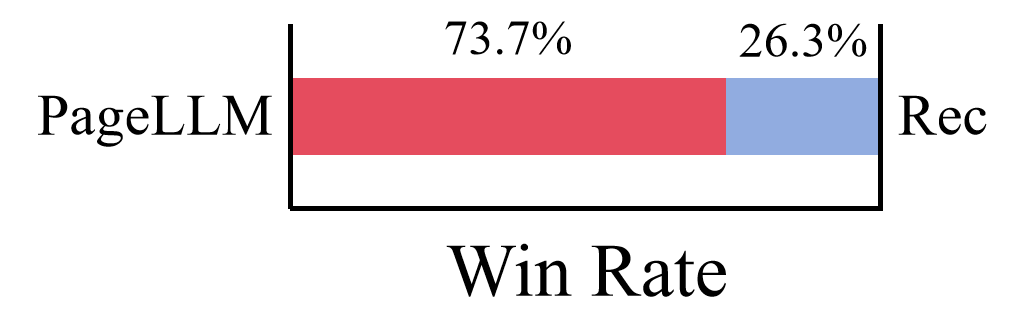}
    \caption{LLM-as-judge pairwise win-rate of PageLLM against its strongest baselines. PageLLM is preferred in the majority of cases on every category.}
    \label{fig:winrate}
\end{figure}

To corroborate these numeric gains with a qualitative signal, we further conduct an LLM-as-judge evaluation: a separate LLM is shown the user prompt and two candidate pages---one from PageLLM and one from a baseline---and asked to pick the more satisfying page. \textbf{Figure~\ref{fig:winrate}} reports the win-rate of PageLLM. PageLLM is preferred in the clear majority of cases on every dataset, aligning qualitative judgment with the quantitative metrics. Because the judge LLM does not have access to user history or ratings at inference, its decisions reflect surface-level indicators (coherence, diversity, relevance to the prompt) rather than personalization. The fact that PageLLM wins on these surface properties as well suggests that the gains are not purely personalization-driven.

\subsection{Cold-Start Robustness (RQ5)}

\begin{figure}[bp]
    \centering
    \includegraphics[width=0.99\linewidth]{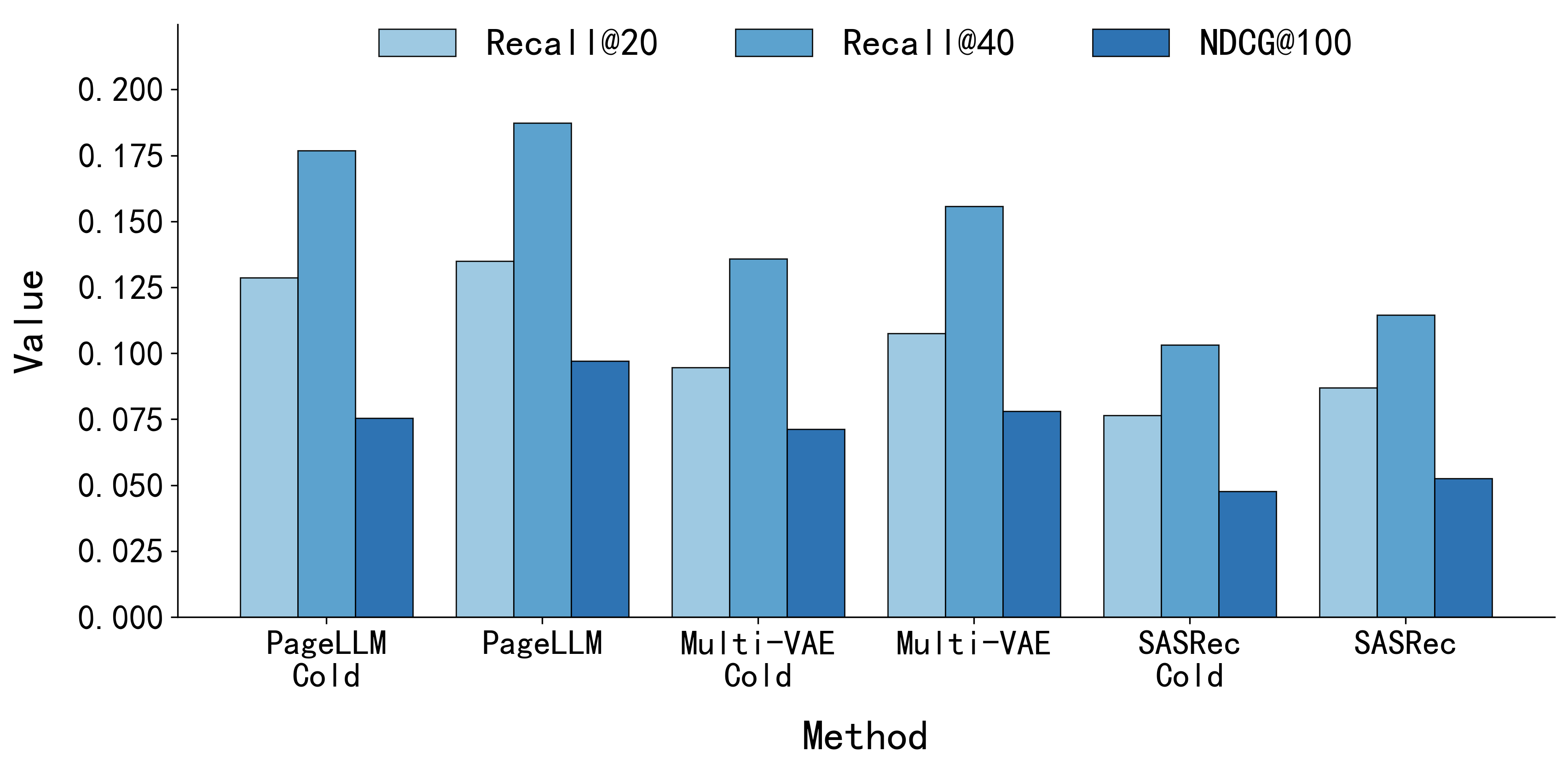}
    \caption{Performance under a 50\% cold-start simulation on AM-Toys. PageLLM degrades less than SASRec and Multi-VAE on every metric, particularly on NDCG@100.}
    \label{fig:cold_start}
\end{figure}

We simulate a data-sparse deployment by training on 50\% of the AM-Toys interactions. \textbf{Figure~\ref{fig:cold_start}} compares PageLLM against Multi-VAE and SASRec, the strongest autoencoder and sequential baselines on this category. Recall@20 drops by only 6.2\% for PageLLM, versus larger relative drops for the two baselines; on NDCG@100, where all methods pay a heavier penalty, PageLLM's decrease (22.3\%) is still the smallest. The pattern suggests that the meta-information pre-training stage endows the model with priors on user--item co-occurrence that survive aggressive subsampling, while the fine-grained reward keeps tail-ranked items well placed even when the supervised signal is diluted.

Two mechanisms jointly explain the cold-start robustness. First, the meta-information pre-training stage uses rating prediction and next-token prediction to transfer knowledge from unaffected user--item relationships, which partially offsets the loss of direct interaction signal. Second, the coarse page-level reward is intrinsically more robust than per-instance supervised loss: it averages over list-level properties (category balance, coverage) that degrade gracefully as data thins out, rather than collapsing when a single positive interaction is removed. Combined, the two mechanisms mean that PageLLM's cold-start degradation is dominated by the item-level reward losing training signal, which is the smallest of the three effects.

\subsection{Backbone Study (RQ6)}

\begin{figure}[t]
    \centering
    \includegraphics[width=0.99\linewidth]{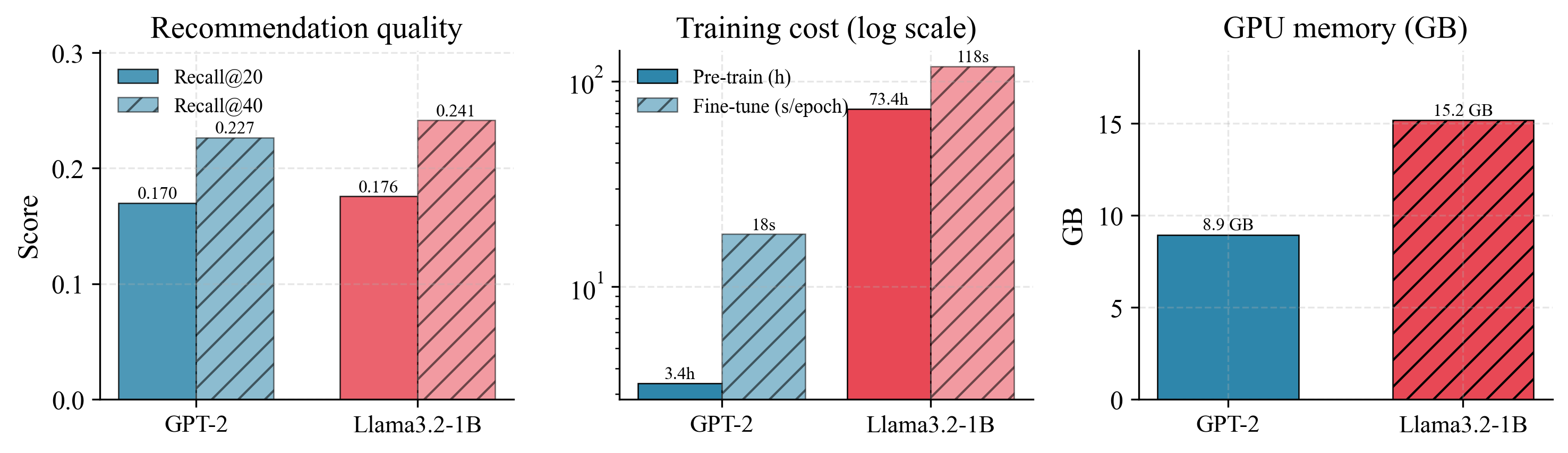}
    \caption{Performance vs.\ cost trade-off between GPT-2 and Llama3.2-1B. Llama yields a small Recall gain at an order-of-magnitude higher pre-training cost; GPT-2 is Pareto-optimal under tight compute budgets.}
    \label{fig:backbone}
\end{figure}

\textbf{Figure~\ref{fig:backbone}} contrasts GPT-2 and Llama3.2-1B as PageLLM's backbone. Llama3.2-1B yields marginally better Recall (0.1757 vs.\ 0.1698 at @20) but at a disproportionately higher training cost: pre-training jumps from 3.4 h to 73.4 h, fine-tuning from 18 s to nearly 2 min per epoch, and peak GPU memory from 8.9 GB (full parameters) to 15.2 GB (LoRA). On the current deployment budget, GPT-2 lies on the Pareto front of quality per dollar; Llama becomes attractive only when the deployment target is large enough for the absolute Recall margin to pay back the extra compute. This trade-off generalizes to other open-source LLM families: in our internal experiments, intermediate-size models such as GPT-2 Large did not change the Pareto boundary materially, which suggests that the limiting factor is the reward signal rather than the backbone capacity.

\subsection{Case Study}

\begin{figure}[b]
    \centering
    \includegraphics[width=0.99\linewidth]{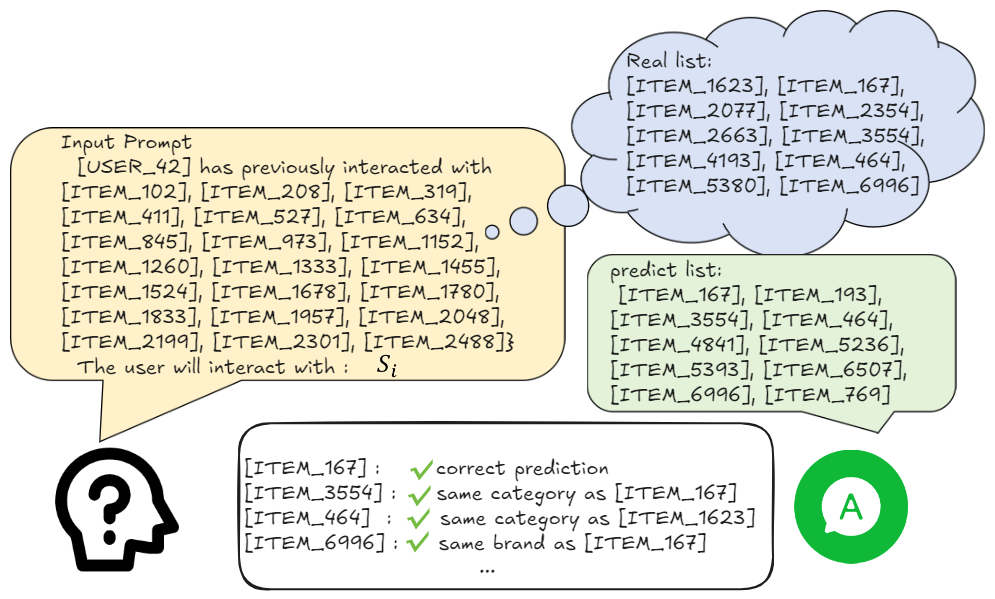}
    \caption{A representative prediction trace. The generated items include an exact match with the ground truth and several semantically related alternatives, illustrating the qualitative effect of the multi-grained reward.}
    \label{fig:case}
\end{figure}

\textbf{Figure~\ref{fig:case}} presents a concrete inference trace. Given a 20-item history for user \texttt{USER\_42}, PageLLM exactly recovers the ground-truth \texttt{ITEM\_167} and surfaces three other items (\texttt{ITEM\_3554}, \texttt{ITEM\_464}, \texttt{ITEM\_6946}) that share category or brand with the target list. The model thus captures both \emph{exact} and \emph{semantic} alignment, which is the operational definition of a good WPO output. Because the multi-grained reward explicitly supervises both list-level structure and per-item placement, the generated list not only contains relevant items but also orders them in a way that resembles how the user historically engaged with similar products---for instance, the high-diversity items appear in the middle of the list rather than clustered at the top, mirroring the pattern in the user's own history.

\subsection{Summary}

Taken together, the six research questions yield a coherent picture. The multi-grained reward improves offline WPO metrics on every Amazon category (RQ1), the improvement survives the transition to production traffic and delivers a significant GMV lift (RQ2), the two reward heads are empirically complementary and cannot be replaced by a single scalar (RQ3), PageLLM outperforms eleven baselines spanning four modeling families (RQ4), the framework degrades gracefully under cold start (RQ5), and its performance--cost trade-off makes GPT-2 the preferable backbone under realistic compute budgets (RQ6). Of these, RQ3 is the load-bearing result: it is the empirical justification for the decoupled reward design that we have been advocating throughout the paper.

%% file: sections/6-conclusion.tex
\section{Conclusion and Future Work}
\label{sec:conclusion}

We have argued that whole-page optimization with large language models is bottlenecked less by the availability of supervision than by the granularity of the supervision signal. PageLLM turns implicit user feedback into four families of contrastive preference pairs and uses them to train a reward with two complementary heads: a coarse page-level reward for list-wide coherence and a fine item-level reward for engagement-sensitive position swaps. The resulting signal drives PPO fine-tuning of a pre-trained LLM without requiring any human annotations.

On seven Amazon categories, PageLLM outperforms eleven strong baselines by as much as 78\% on NDCG@100; a controlled ablation confirms that each reward head is necessary and that their combination is super-additive; a 10M-user online A/B test yields a 0.44\% lift in GMV and consistent up-funnel gains; and cold-start, backbone, and LLM-as-judge studies corroborate the framework's robustness. The additive combination of the two reward heads also enables incremental deployment: the coarse reward can be served as a CPU-side ranking feature before the full LLM policy is rolled out, which lowers the operational barrier in our own production experiment.

Several directions remain open. First, the four preference-pair families we use are a deliberate but not exhaustive taxonomy of page-level failure modes; adding pairs that capture modality-specific failures (e.g., mismatched images, stale prices) is a natural extension once multimodal LLM backbones become more efficient to fine-tune. Second, the item-level reward currently ignores cross-list interactions; a richer reward that models how items on one page influence future sessions could tighten the online lift further. Third, the framework assumes a stationary catalog; extending it to cross-domain and continually changing catalogs---where new users and items arrive faster than the reward model can be retrained---is both practically important and technically non-trivial. We view these as promising follow-up questions, rather than fundamental limitations of the multi-grained reward idea that PageLLM introduces.